\documentclass[letterpaper, 10 pt, conference]{ieeeconf}

\IEEEoverridecommandlockouts

\usepackage[pdftex]{graphicx}
\usepackage{amssymb,amsmath,url}
\usepackage{bm} 
\usepackage{multirow}
\usepackage{mathtools}
\usepackage[dvipsnames]{xcolor}
\usepackage{booktabs}	
\usepackage{epstopdf}
\usepackage{dsfont}
\usepackage{caption}
\usepackage{float}
\usepackage{bigfoot}		
\usepackage{algpseudocode}
\usepackage{pifont}
\usepackage{tikz}
\usepackage{comment}
\usepackage{cite}
\usepackage{lipsum}
\usepackage{array}
\usepackage{pdfpages}
\usepackage[titlenumbered,ruled,lined,linesnumbered]{algorithm2e}
\usepackage{setspace}
\usepackage{booktabs} 
\usepackage{rotating} 
\usepackage{array} 
\usepackage{algpseudocode} 
\usepackage{titlesec}
\usepackage{hyperref}

\allowdisplaybreaks

\algrenewcommand\algorithmicrequire{\textbf{Input:}}
\algrenewcommand\algorithmicensure{\textbf{Output:}}
\algrenewcommand\algorithmicforall{\textbf{For}}

\newtheorem{theorem}{Theorem}

\newtheorem{proposition}{Proposition}

\allowdisplaybreaks
\newtheorem{definition}{Definition} 

\newtheorem{assumption}{Assumption}
\newcounter{relctr} 
\everydisplay\expandafter{\the\everydisplay\setcounter{relctr}{0}} 

\AtBeginDocument{}

\newcommand{\R}{\mathbb{R}}

\newcommand{\N}{\mathbb{N}}

\DeclareMathOperator*{\argmax}{arg\,max}
\def\revi#1{\begingroup   #1\endgroup}
\def\ques#1{\begingroup \color{green}   #1\endgroup}

\newcommand{\hsp}{\hspace{2mm}}

\let\oldnl\nl
\newcommand{\nonl}{\renewcommand{\nl}{\let\nl\oldnl}}

\usepackage{setspace}

\graphicspath{{images/}{../images/}}

\title{\LARGE \bf \revi{Multi-robot task allocation for safe planning\\ against stochastic hazard dynamics}
}

\author{Daniel Tihanyi, Yimeng Lu, Orcun Karaca, and Maryam Kamgarpour
\thanks{\textit{Corresponding author: M. Kamgarpour}.\newline\indent Tihanyi, Lu are with ETH Z\"{u}rich, Switzerland. emails: {\tt \{tihanyid, luyi\}@ethz.ch}, Karaca is with ABB Corporate Research, Switzerland. email: {\tt orcun.karaca@ch.abb.com}.  Kamgarpour is with Systems Control and Multiagent Optimization Research (sycamore) lab at EPFL, Switzerland. email: {\tt maryam.kamgarpour@epfl.ch}.} }

\allowdisplaybreaks
\begin{document}

\maketitle

\begin{abstract}
We address multi-robot safe mission planning in uncertain dynamic environments. This problem arises in several applications including safety-critical exploration, surveillance, and emergency rescue missions. Computation of a multi-robot optimal control policy is challenging not only because of the complexity of incorporating dynamic uncertainties while planning, but also because of the exponential growth in problem size as a function of number of robots. Leveraging recent works obtaining a tractable safety maximizing plan for a single robot, we propose a scalable two-stage framework to solve the problem at hand. Specifically, the problem is split into a low-level single-agent control problem and a high-level task allocation problem. The low-level problem uses an efficient approximation of stochastic reachability for a Markov decision process to derive the optimal control policy under dynamic uncertainty.
The task allocation is solved using polynomial-time forward and reverse greedy heuristics and in a distributed auction-based manner. By leveraging the properties of our safety objective function, we provide provable performance bounds on the safety of the approximate solutions proposed by these two heuristics. We evaluate the theory with extensive numerical case studies.
\end{abstract}


\vspace{-.4cm}
\section{Introduction}
\vspace{-.1cm}

Autonomous robots are increasingly used in safety-critical applications including surveillance \cite{di2010autonomous, jorgensen2017matroid,zhou2018resilient} and emergency rescue missions \cite{wood2013stochastic,wood2016automaton,lu2020safe}. Safety against dynamic uncertainties, such as moving obstacles and evolving hazards is indispensable in such applications\cite{herbert2017fastrack,manjunath2021safe,sahraeekhanghah2021pa,warnke2020towards}. A natural idea to improve safety is to use multiple robots. This setup is commonly used in situations where the workload can be distributed to the robots to reduce the execution time by working in parallel~\cite{yan2013survey, fazlollahtabar2015methodologies,khoo2019distributed}, and to increase robustness due to redundancy~\cite{khamis2015multi}.

In an emergency rescue scenario, the objective of visiting a set of target locations (e.g., to save survivors) can be fulfilled collectively by a team of robots, where each robot is assigned a safety-maximizing trajectory visiting a subset of these targets. This objective can be formulated as a multi-robot optimal control synthesis problem for a Markov decision process. The challenge in solving this problem is twofold. The first comes from computing a target-robot assignment which maximizes the safety of all robots while visiting all target locations. The second comes from tractably solving for the safety-maximizing trajectories of the robots given their assigned targets. {Past works provide methods based on dynamic programming to solve the latter problem through the framework of probabilistic safety and reachability ~\cite{lu2020safe,wood2013stochastic, zhou2018joint, smith2011optimal,wolff2012robust}}. Leveraging an efficient implementation from these methods, our work provides a scalable two-stage framework for solving the multi-robot task/target allocation problem to maximize the safety of the mission.\looseness=-1

The number of all possible task assignments to consider grows exponentially with the number of both robots and targets~\cite{korsah2013comprehensive}. Moreover, such task allocation problems generalize the well-studied set partitioning problem, which is  NP-hard~\cite{wolsey1999integer,gerkey2004formal}. {Having a suboptimal task assignment and not accounting for safety objectives would jeopardize safety of the mission. Thus, it is desirable to formalize the safety objective in task allocation and derive guarantees on the performance/safety of the task allocation algorithms.}
\revi{Auction-based approaches~\cite{schillinger2018auctioning, faruq2018simultaneous, lagoudakis2005auction,berhault2003robot} consider the robots as bidders who iteratively submit bids for the most desirable tasks. Such iterative and distributed assignments of tasks are instances of the efficient forward greedy heuristics, which could provide suboptimality guarantees~\cite{nemhauser1978analysis}.}\footnote{The forward greedy refers to the greeedy heuristic that adds elements iteratively, whereas the reverse greedy refers to the one that removes.} {In contrast with the existing auction-based approaches, our objective function is a safety measure defined by the completion of the tasks for each robot. This function originates from the underlying low-level stochastic optimal control problem~\cite{wood2016automaton}. The collective goal of multi-robot planning is to maximize such a safety objective, and in our case, we propose a {multiplicative form that allows a distributed implementation of greedy heuristics. Hence, our proposed methods are variations on the auction-based approaches, in which the safety objective is incorporated into the bids and allocations.}}\looseness=-1

\revi{Greedy algorithms are equipped with provable performance bounds when the objective functions satisfy sub- or supermodularity assumptions~\cite{nemhauser1978analysis}. However, we will show that our safety objective is both nonsubmodular and nonsupermodular.
The aforementioned studies on auction-based approaches either do not mention/propose any optimality guarantee, or when they do, their problem formulations do not capture nonsub- and nonsupermodular objective functions. Keeping this in mind, we provide a literature review on both the auction-based approaches and general set partitioning studies. In the studies of \cite{khoo2020distributed} and also in the case of the distributed Hungarian algorithm of \cite{chopra2017distributed}, a set partitioning problem  is formulated. This problem is based on a bipartite graph requiring fixed target-robot pair costs for edges (that is, additive/modular objective functions). Our nonmodular objective associates costs to subset of all tasks allocated to robots, which makes these methods inapplicable to our problem. In contrast, other methods such as \cite{atamturk1996combined} formulate a set partitioning problem while allowing costs to be associated with subsets of all tasks. However, they neglect possible differences between individual robots and their objective functions, moreover, they require an exhaustive list of all costs for all subsets. In our case, different robots might succeed with different probabilities when assigned the same set of tasks, and this aspect cannot be captured by \cite{atamturk1996combined}. Other methods in the literature for task allocation problems assume submodular objective functions~\cite{li2019threshold,lagoudakis2005auction}. Specifically, implementing descending price iterative auction algorithm of \cite{li2019threshold} but with a nonsubmodular objective can lead to nonconvergence and/or no performance guarantee. In terms of safety-oriented task allocation, there are studies where the objective is either the conditional value-at-risk cost~\cite{nam2016analyzing, yang2017algorithm}, or the worst-case cost~\cite{ponda2012distributed}. Objectives in these works are also additive.
To the best of our knowledge, general nonsub- and nonsupermodular objective functions have not been addressed in any of the existing task allocation studies or in related applications of set partitioning problems.}\looseness=-1

\revi{We show through numerical studies that the our safety objective is weakly sub- and supermodular, notions characterized by submodularity ratio and curvature, respectively. Thus, we  leverage recent theoretical results from \cite{guo2019actuator, guo2020actuator,orcun2021performance} to obtain safety guarantees on our auction-based algorithms. 
To the best of our knowledge, this work is the first to demonstrate the benefits of a novel auction-based task allocation using the reverse greedy algorithm both in theory and in numerics. Past auction-based methods considered the forward greedy algorithm, and we show that the reverse has a better theoretical guarantee than the forward for a large set of parameters in our problem framework. }

Our contributions are summarized as follows. 

(i) We develop a scalable two-stage framework for an emergency rescue scenario by splitting the multi-robot controller synthesis problem into a safe planning problem (for each robot) and a multi-robot task allocation. To this end, we utilize an efficient implementation of a single-robot plan under dynamic uncertainties. This approach serves as the low-level planner, and it allows the computation of the multiplicative safety objective for the high-level task allocation. \looseness=-1

(ii) \revi{To allocate the targets in a tractable manner, we introduce two variants of greedy algorithms, the forward and the reverse. We show that the multiplicative safety  formulation of the objective decouples the individual robots’ optimal control problems under a fixed task assignment. This enables applying these greedy algorithms in a distributed auction-based manner. Here, the agents submit bids based on their individual objectives, and a central unit (or one of the agents) chooses the best among these bids based on the group objective. Moreover, under  weak sub- and supermodularity properties we provide performance guarantees on the safety of the forward and the reverse greedy solutions.} 

(iii) We compare these two greedy algorithms  in terms of their theoretical guarantees, and computational and practical performance in numerical case studies. Theoretical analyses suggest that reverse greedy algorithm can have a better guarantee than the forward greedy algorithm in a larger range of problem instances. However, this improved theoretical guarantee of the reverse greedy comes with an increased computational complexity. 
In terms of empirical performance, we observe that both algorithms perform similar and close to optimality, when compared to the brute force optimal solution.
Our case studies are based on the implementation of our two-stage framework in example environments for an emergency rescue scenario. \revi{Our code is publicly available at {\href{http://github.com/TihanyiD/multi_alloc}{\textcolor{blue}{github.com/TihanyiD/multi\_alloc}}}.}

\textbf{Organization}. \revi{Description and modeling of the environment and the overall problem statement are presented in Section~\ref{sec:modelling}.}
 We then provide our two-stage framework by formulating both the single-robot safe planning and the multi-robot task allocation problems in Section~\ref{sec:framework}.
 Section~\ref{sec:greedy} proposes the two distributed greedy approaches for task allocation. Finally, we study a numerical case study comparing algorithms and their performance
 in Section~\ref{sec:case_studies}. To support our conclusions, we provide in an appendix additional extensive case studies, including randomized instances.


\section{Problem formulation and statement}\label{sec:modelling}

Consider a team of autonomous robots operating in an environment containing obstacles (e.g., walls), a set of targets (e.g., survivors), and a stochastically evolving hazard (e.g., fire or toxic contamination). The goal of the robots is to visit the targets and exit while avoiding unsafe hazard locations. We will model the problem in detail and formulate it as a stochastic reachability problem as follows. 

\subsection{Modeling the environment and the robots}
\textbf{Assumptions.} We assume that the map of the environment including the obstacles and the location of the targets and hazard sources are known a priori. Furthermore, we can describe the hazard spread with a known stochastic model. The map consists of cells arranged in a discrete grid. Collision avoidance among the robots is outside of the scope of our current study. Hence, we assume grid cells are large enough so that multiple robots can occupy a cell simultaneously.

\begin{figure}[t]
    \centering
    \includegraphics[width=0.50\linewidth]{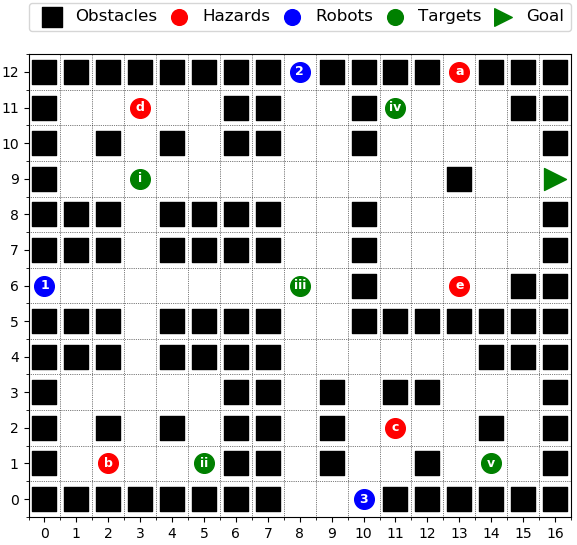}
    \caption{An example setup of our problem. The robots need to collectively visit all the targets and reach the goal while avoiding the evolving hazard.}\label{fig:map_1}\vspace{-.1cm}
\end{figure}

\textbf{Map model.} The environment is modeled by a grid map as illustrated with an example in Figure~\ref{fig:map_1}. We let $M_{m \times n}$ be an $m \times n$-sized map and $O \subset M_{m \times n}$ be a set of obstacles. $X=M_{m \times n}\setminus O$ is the set of free cells. For the  cell $x \in X$, let $N(x)$ be the set of neighboring cells of~$x$. Such a grid-based map has been introduced by \cite{elfes1987sonar}, and is common in the robotics literature to model free space and obstacles \cite{hornung2013octomap}.\looseness=-1

\textbf{Robots and targets.} Define the set of robots as $R=\{1,\ldots,\rvert R\rvert\}$ and the target set as $T \subset X$.

\textbf{Robot motion.} The set of possible inputs correspond to the direction the robot can move: $U=\allowbreak\{\text{Stay},\allowbreak\,\text{North},\allowbreak\,\text{East},\allowbreak\,\text{South},\allowbreak\,\text{West}\}$. Thus,
    $d_\text{Stay}=(0,0),\ d_\text{North}=(0,1),\  d_\text{East}=(1,0),\  d_\text{South}=(0,-1),\ d_\text{West}=(-1,0)$.
In each position $x \in X$, the set of 
    $U(x)=\left\{u \in U\,\rvert\,x+d_u \in X\right\} \subseteq U,$
are the inputs available to the robot. The motion of the robot is defined by a stochastic transition kernel $x^{k+1} \sim \tau_X\left(\cdot\,\rvert\,x^k,u^k\right)$, $k \in \{0,1,\dots\}$ with initial position $x^0\in X$, where $\tau_X: X \times X \times U \rightarrow [0,1]$ denotes the probability of transiting from $x^k \in X$ at time step $k$ to $x^{k+1} \in N(x)$ at  $k+1$ under control input $u^k \in U(x^k)$. The stochastic model accounts for potential uncertainty of the robot control.\footnote{A deterministic robot dynamic can be considered by defining $\tau_X$ as\looseness=-1
\begin{equation*}
    \tau_X\left(x^{k+1}\,\rvert\,x^k,u^k\right)=
    \begin{cases}
        1 & \text{if} \hsp x^{k+1}=x^k+d_{u^k},\\
        0 & \text{otherwise}.
    \end{cases}
\end{equation*}}

\textbf{Hazard spread.} Let $Y=2^X$ be the hazard state space. Each element $y \in Y$ denotes a set of hazardous cells. The stochastic Markov process $y^{k+1} \sim \tau_Y\left(\cdot \,\rvert\,y^k\right)$, $k \in \{0,1,\dots\}$ defines the hazard evolution dynamics with transition kernel $\tau_Y: Y \times Y \rightarrow [0,1]$ from state $y^k \in Y$ at time step $k$ to $y^{k+1} \in Y$ at time step $k+1$. Detailed dynamics of the transition kernel depends on the modeling assumptions. For example, in the case of fire, an estimation of the probability of fire spread from a given grid to the neighboring grids can be used to derive $\tau_Y$ from~\cite{beer1990fire,tymstra2010development,wood2016automaton}. Details on our specific modelling choice for the numerical studies are provided in Appendix~\ref{sec:apx_tau_Y_model}.

\subsection{Problem statement for multi-robot multi-task safe control}\label{sec:problem}
The \textit{mission} of the robot fleet is to visit every target and exit the hazard site safely (e.g., a rescue scenario). Implementing safety as a hard constraint by considering the worst-case hazard spread generally yields infeasibility. To this end, our objective is instead to determine each robot's control input at every time step to maximize the probability of completing the mission within a given time horizon $N \in \N_{>0}$ while avoiding the stochastically evolving hazard. By considering the robot fleet as one system, whose state-space is the product space of each robot's state-space $X^{|R|}$, this objective can be formalized as a stochastic reachability problem for a suitably defined Markov decision process \cite{abate2008probabilistic,summers2013stochastic,lu2020safe,wood2016automaton,bansal2017hamilton}. The stochastic reachability problem can then be solved using the dynamic programming principle as shown in the above works. We postpone the precise mathematical definition a single-robot formulation of this problem to the next section.\looseness=-1

Given that maximizing the mission success probability relies on dynamic programming on the robots' product space, the computation of a multi-robot optimal control policy is intractable. This motivates a two-stage approach to compute an approximate solution, as detailed in the following section.

\vspace{-.1cm}

\section{Two-stage multi-robot safe planning}\label{sec:framework}
We present a scalable framework to allocate targets to robots and control the robot fleet to maximize safety. We split the problem into the following two stages: low-level path planning (optimizing control policies of each robot individually for assigned targets) and high-level task allocation. 

\subsection{Single-robot path planning}\label{sec:low_level}
Leveraging the works of \cite{wood2013stochastic,wood2016automaton,lu2020safe}, we provide a tractable solution to the safe control problem for a single-robot system and a given set of targets. This serves as a building block of our multi-robot framework. To this end, we first define a stochastic Markov process combining the model of the robot's motion, its knowledge of the hazard spread, and its progress towards completing its mission. We build on our definitions in Section~\ref{sec:modelling} describing the behaviour of the environment. We then utilize a dynamic programming algorithm to find the optimal control policy by maximizing the probability of success. 

\subsubsection{Definition of the stochastic process} 
The discrete-time stochastic process is defined by its state space and transition kernels as follows.

\textbf{State space.} \textit{Robot state} -- At each time step $k$ we keep track of the 2D position of the robot $x^k \in X$ on the map.

\textit{Task execution} -- During the execution of the mission, the robot needs to keep track of the visited targets. We first define $T_r \subset T \subset X$ as the \textit{target list}, the set of targets assigned to robot $r$. We then define the set $Q=2^{T_r}$ and the \textit{target execution state} $q^k \in Q$ at time step $k$. The state~$q^k$ tracks the visited targets.

\textit{Hazard state} -- As discussed in the previous section, the state of the hazard is $y^k \in Y$, the set of grid cells that are hazardous at time step $k$. Including the hazard state $y^k$ in the state space results in a computationally intractable problem.\footnote{The cardinality of $X$ grows with the map's size, whereas the number of target execution states $|Q|=2^{|T_r|}$ grows exponentially with the size of $T_r$. In practice, we have $|X|\gg|T_r|$. Thus, the hazard state space's size $|Y|=2^{|X|}$ would be the main source of complexity.} To address this issue, a single \textit{hazard state} denoted by $s_H$ is introduced to capture the case where the robot enters a hazardous cell, namely, $x^k \in y^k$. A robot reaching the hazard state indicates an \textit{unsuccessful mission} and the robot will remain in this state.\looseness=-1

Combining the elements mentioned above, we define the state space 
$S=\{s_H\} \cup (Q \times X\setminus \{(q,x)\,\rvert\, x \in T_r \wedge x \notin q\}) .$ As we take a union with the single hazard state object defined above, we obtain a reduction in size when compared to including the full state of the hazard.
We specify the \textit{goal location} as $x_G \in X$. This refers to the exit of the map. We also define the \textit{goal state} denoted by $s_G=(T_r,x_G)$. The state $s_G$ indicates a \textit{successful mission}, where every target is visited and the robot reaches the safe goal location. Finally, we define the initial state for robot $r$ as $s_r^0=(\emptyset,x_r^0)$, where no targets are visited and the robot is at its initial position $x_r^0$. The state $s_r^0$ is known to the robot.\looseness=-1

\textbf{Transition probabilities of the stochastic process.} Let $\tau_S^k: S \times S \times U \rightarrow [0,1]$ be the transition kernel at time step $k$. The quantity $\tau_S^k(s^{k+1}\,\vert\,s^k,u^k)$ represents the probability of a state transiting from $s^k$ to $s^{k+1}$ given the control input $u^k$ at time $k$. This kernel can be computed using the robot dynamics ($\tau_X$), the execution of the tasks, and the uncertain hazard dynamics ($\tau_Y$). We provide the detailed mathematical description in Appendix~\ref{sec:apx_tau_S_k}.

\subsubsection{Controller synthesis via dynamic programming}
\label{subsub:controller}

Let $\pi=\{\mu^0,\dots,\mu^{N-1}\}$ be a  control policy, where $\mu^k: S \rightarrow U$ refers to the control law at time step $k$. With the state-space defined above, the objective of mission success defined in Section \ref{sec:problem} can be cast as a stochastic reachability problem: finding a control policy so that the probability of reaching the goal state $s_G$ within a given time horizon $N \in \N_{>0}$ is maximized. Parameter $N$ is chosen to be sufficiently long to ensure the robot can reach the targets. 

For the remainder, $f_r(T_r)$ denotes the probability of reaching the goal state of robot $r$ under the optimal control policy $\pi_r(T_r)$ for a given target list $T_r$, whereas $f_r(\pi,T_r)$ denotes the probability of success under the control policy~$\pi$.  Moreover, we will use the terms success and safety probabilities interchangeably, and both will imply the execution of the assigned tasks by the robot (or the robot fleet).
Given these definitions, our goal is to solve
\begin{equation}\label{eq:optimal_policy}
    \pi_r(T_r)=\argmax_{\pi} f_r(\pi,T_r).
\end{equation}
Problem~\eqref{eq:optimal_policy} of a single robot can be solved using Algorithm~\ref{alg:dynamic_programming_algorithm}\cite{abate2008probabilistic,summers2011stochastic} in a tractable way, similar to the dynamic programming approach to a stochastic control problem. The function value $V^k(s)$ captures the maximum probability of reaching the goal state starting from time step $k$ at state $s$, and is defined recursively in the algorithm. It can be shown that $f_r(T_r)=\max_{\pi} f_r(\pi, T_r)=V^0(s_r^0)$, whereas $\pi_r(T_r)$ can be computed as the input achieving the maximum in line 5 of the algorithm above. It can easily be verified that $f_r(\cdot)$ is nonincreasing as a function of set of tasks.


\begin{algorithm}[t]\vspace{.1cm}
    \setstretch{.7}
    \setcounter{AlgoLine}{0}
	\KwIn{robot $r$, horizon $N \in \N_{>0}$, targets $T_r$} \KwOut{policy $\pi_r(T_r)$, probability of safety $f_r(T_r)$}
	\SetKwBlock{Begin}{begin}{end}
	\Begin{
		initialization: $V^N(s)=
		    1\  \text{if} \hsp s=s_G;\
            0\  \text{otherwise}$\\
		\For{$k=N-1,\dots,0$}{
		    $V^{k-1}(s)=\max\limits_{u \in U(x^{k-1})}\big\{\sum\limits_{s' \in S}\tau_S^{k-1}(s'\,\vert\,s,u)\cdot V^k(s') \big\}$
		}
	}\caption{Dynamic Programming Algorithm}\label{alg:dynamic_programming_algorithm}
\end{algorithm}

\subsection{Multi-robot task allocation}\label{sec:high_level}
The high-level stage of our framework assigns the targets to the robots to maximize a collective objective. This part builds on the solutions obtained by the low-level stage described above. We start by formulating the task allocation problem and then introduce the collective objective function. 

 We assume it is sufficient that a target is visited once. Thus, a feasible task allocation assigns each target to exactly one robot: partitioning the set $T$ into $\{T_r\}_{r \in R}$. To be more specific, $T_r \subset T$ for all $r \in R$, $T_r \cap T_{r'}=\emptyset$ for any pair $r,r' \in R$ where $r\neq r'$ and $\bigcup_{r \in R}T_r=T$.

\textbf{Objective function.}
The goal is to find a feasible task allocation that maximizes the \textit{probability of group safety}, that is, the probability of all robots safely finishing their missions (assigned tasks) simultaneously.
Denote this probability by $f_R(\{T_r,\pi_r\}_{r \in R})$ for a fixed set of assigned tasks and control policies, and its optimizer by $\{T_r^\star,\pi_r^\star\}_{r \in R}$. However, maximizing the probability of group safety is computationally challenging since it requiries formulating the stochastic reachability of Section \ref{subsub:controller} over a product state space $X^{|R|}$ considering the whole robot fleet in a centralized manner. Thus, we introduce the \textit{multiplicative group safety} as an approximation of this objective function. It is defined as $
    F(\{T_r\}_{r \in R})=\prod_{r \in R} f_r(T_r),$
where the values of $f_r(T_r)$ are obtained by solving the single-robot path planning problem introduced in Section~\ref{sec:low_level}. 

Note that the robots use the same map under the same hazard state evolution, thus, the safety of individual robots are not independent events. This implies that the probability of group safety can differ from the multiplicative group safety. 
{To support this objective function choice, we now show that the multiplicative group safety is a lower bound to the probability of group safety under a mild assumption. Specifically, the assumption requires that 
conditioning on the success/safety of other robots does not decrease the probability of a robot successfully completing its own set of  tasks. 
\looseness=-1

\begin{assumption}\label{assm:1}
Consider a fixed task allocation and a control policy for each robot. Let $E_r$ denote the random variable with support $\{0,1\}$ defining the safety of robot $r$ by taking the value $1$ with probability $P(E_r=1)$. We assume $P(E_r=1\rvert \prod_{r'\in R'} E_{r'}=1 )\geq P(E_r=1)$ for any $r\in R$ and $R'\subset R$.
\end{assumption}

In realistic examples, the  assumption above holds  since an observation regarding the safety of other robots confirms that the hazard did not propagate in certain regions of the map. We remind the reader that the grid cells are assumed to be large enough to not be concerned with collision avoidance, which may otherwise jeopardize this assumption.}
In a more general setting, our condition in Assumption~\ref{assm:1} is also implied whenever knowing that a cell is safe at any time point does not decrease the safety probability of any other cell at any future time. This is also a reasonable core assumption and can be verified numerically when considering hazard models describing fire in \cite{beer1990fire,tymstra2010development}.

\revi{\begin{proposition}
Under Assumption~\ref{assm:1}, multiplicative group safety is a lower bound to the probability of group safety: $f_R(\{T_r^\star,\pi_r^\star\}_{r \in R})\geq f_R(\{T_r,\pi_r(T_r)\}_{r \in R}) \geq F(\{T_r\}_{r \in R})$.
\end{proposition}
\begin{proof}
Observe that $P(\prod_r E_r=1)=P(E_1=1,\ldots,E_{|R|}=1)=\prod_r P(E_r=1\rvert \prod_{r'=1}^{r-1}E_{r'}=1)$ via chain rule. Clearly, the condition in the proposition above yields $\prod_r P(E_r=1\rvert \prod_{r'=1}^{r-1}E_{r'}=1)\geq \prod_r P( E_r=1)$. Thus, we obtain $P(\prod_r E_r=1)\geq \prod_r P( E_r=1)$. Under a change of notation, this is equivalent to $f_R(\{T_r,\pi_r(T_r)\}_{r \in R}) \geq \prod_{r \in R} f_r(T_r)= F(\{T_r\}_{r \in R}).$
The remaining inequality in the proposition statement, $f_R(\{T_r^\star,\pi_r^\star\}_{r \in R})\geq f_R(\{T_r,\pi_r(T_r)\}_{r \in R})$, follows from the optimality of task assignments and policies listed in $\{T_r^\star,\pi_r^\star\}_{r \in R}$. This concludes the proof.
\end{proof}}
The above results implies that the value of the introduced multiplicative objective function at a given policy implies a lower bound on the probability of group safety of the policy.

For the remainder, we consider the multiplicative group safety to be our success/safety measure.

\textbf{Optimization problem.} The task allocation problem for maximizing safety can thus be formulated as
\begin{align}\label{eq:task_allocation_problem}
	F^*=&\max_{\{T_r\}_{r \in R}}\ \prod_{r \in R} f_r(T_r)\\ &\quad\mathrm{ s.t. }\quad T_r \cap T_{r'}=\emptyset \text{, } \forall r\neq r' \text{, } \cup_{r \in R} T_r=T.\nonumber
\end{align}

The problem above generalizes set partitioning, which is NP-hard \cite{wolsey1999integer,gerkey2004formal,shin2010uav}. We highlight that none of the past works we reviewed in the introduction had a safety objective as in \eqref{eq:task_allocation_problem}. In the next section, we propose two variants of the greedy algorithms to enable efficient distributed implementations of task allocation for maximizing safety.

\section{Greedy heuristics}\label{sec:greedy}
We introduce two greedy heuristics, the forward and the reverse, in Sections~\ref{sec:forward} and~\ref{sec:reverse}, respectively. From the practical standpoint, these algorithms iteratively update the task allocation by adding or removing one task-robot pair at each step. Moreover, thanks to its multiplicative form, our objective can be implemented in an auction-based fashion. At each iteration, each robot can submit a bid based on their individual objective $f_r$. The bids are then collected by a central unit such that a decision based on the group objective $F$ can be made. The properties of the group objective $F$ and theoretical performance guarantees of the algorithms are discussed in Section~\ref{sec:guarantees}.\looseness=-1
\vspace{-.1cm}
\subsection{Forward greedy algorithm}\label{sec:forward}
The forward greedy algorithm (see Algorithm~\ref{alg:fg_distributed}) is initialized with no tasks allocated to any of the robots. It then iteratively updates this allocation by choosing the task-robot pair with the best optimality gain until every task is allocated to one robot. The computation of the iteration steps is distributed among the robots. 

\begin{algorithm}[t]
    \setstretch{0.95}
    \setcounter{AlgoLine}{0}
	\KwIn{$R$, $T$, $\{f_r\}_{r \in R}$} \KwOut {$\{T_r^\text{fg}=T_r^{|T|}\}_{r \in R}$}
	\SetKwBlock{Begin}{begin}{end}
	\Begin{
		initialization: \mbox{$T_r^0=\emptyset$, $f_r^0=f_r(\emptyset)$, $\forall r$, $J^0=T$, $R^0=R$}\label{alg:fg_distributed_line_1}\\
		\For{$k=1,\dots,|T|$}{ \label{alg:fg_distributed_line_2}
			\For{$r \in R^{k-1}$}{\label{alg:fg_distributed_line_3}
				\mbox{$t_r^k \gets \argmax\limits_{t \in J^{k-1}} f_r\left(T_r^{k-1}\cup t\right)-f_r\left(T_r^{k-1}\right)$}\\
				
				\mbox{$\delta_r^k \gets f_r\left(T_r^{k-1}\cup t_r^k\right)-f_r\left(T_r^{k-1}\right)$}\label{alg:fg_distributed_line_4}
			}
			\mbox{$(t_r^k,\delta_r^k) \gets (t_r^{k-1},\delta_r^{k-1}) \hspace{2mm} \forall r \notin R^{k-1}$}\label{alg:fg_distributed_line_5}\\
			
			\mbox{$r^k \gets \argmax_{r \in R} \delta_r^k \cdot \prod_{r' \in R\setminus\{r\}} f_{r'}^{k-1}$}\label{alg:fg_distributed_line_6}\\
			
			\mbox{$f_r^k \gets 
					f_r^k-\delta_r^k \text{, if } r=r^k;\
					f_r^{k-1} \text{ otherwise}$}\label{alg:fg_distributed_line_7}\\
				
			\mbox{$	T_r^k \gets 
    				T_r^k \cup t_r^k \text{, if } r=r^k;\
    				T_r^{k-1} \text{, otherwise}$}\\
				
			\mbox{$R^k \gets \left\{r \,\rvert\,t_r^k=t_{r^k}^k\right\}$}\label{alg:fg_distributed_line_9}\\
			
			\mbox{$J^k \gets J^{k-1} \setminus t_{r^k}^k$}\label{alg:fg_distributed_line_8}
		}
		$\{T_r^\text{fg}=T_r^{|T|}\}_{r \in R}$
	}
	\caption{Forward Distributed Greedy Algorithm}\label{alg:fg_distributed}
\end{algorithm}

Algorithm~\ref{alg:fg_distributed} proceeds as follows. First, define the following variables for each step $k$: $\{T_r^k\}_{r \in R}$ denotes the current task allocation, and $\{f_r^k\}_{r \in R}$ stores the evaluated safety objective function values for each robot $r$. Furthermore, $J^k$ is the set of tasks yet to be allocated and $R^k$ is the set of robots which needs to update their bids in the next step. Initially, no tasks are assigned, hence $T_r^0=\emptyset$ for all $r \in R$ (see Line~\ref{alg:fg_distributed_line_1}). In each step exactly one task is allocated, hence we need $|T|$ steps to complete the algorithm (Line~\ref{alg:fg_distributed_line_2}). At each iteration $k$, all robots $r \in R$ submit a bid (see Line~\ref{alg:fg_distributed_line_3}--\ref{alg:fg_distributed_line_5}), which consists of the pair $(t_r^k,\delta_r^k)$. Each robot $r$ chooses the task $t_r^k$ from the list of unallocated tasks $J^{k-1}$, such that it obtains the best optimality gain $\delta_r^k$ with respect to the individual objective function $f_r$. After collecting all the bids, we choose the robot $r^k$ which generates the best optimality gain with respect to the collective objective: the multiplicative group safety $F$ (Line~\ref{alg:fg_distributed_line_6}). Due to our auction-based formulation in this line, we can choose the task-robot pair with the best collective gain efficiently. Between Lines~\ref{alg:fg_distributed_line_7}--\ref{alg:fg_distributed_line_8}, we simply set the values of $f_r^k$, $T_r^k$ for all $r \in R$ and $R^k$, $J^k$ according to our choice from the task allocation.\footnote{Note that only the robots choosing the same task as $r^k$ \revi{(i.e., $r$ such that $t_r^k=t_{r^k}^k$)} have to update their bids in the next iteration (see Line~\ref{alg:fg_distributed_line_9} at step $k$ and Line~\ref{alg:fg_distributed_line_3} at step $k+1$). The rest of the robots simply submit their bids from the previous iteration (see Line~\ref{alg:fg_distributed_line_5}). }
As a remark, one can also implement the greedy algorithm using a log transform of our objective yielding instead the objective $\sum_{r \in R} \log (f_r(T_r))$. This transformation would still yield the exact same algorithm steps and solutions for both the forward and later the reverse greedy algorithms.\footnote{Moreover, the function $\sum_{r \in R} \log (f_r(T_r))$ is again both nonsubmodular and nonsupermodular with rescaled ratios for the performance guarantees in the next subsection. Submodular/supermodular like properties are well-established to be invariant under strictly increasing reparameterizations (which includes log-transform)\cite{bach2019submodular}.}

We will later see that the solution obtained by this algorithm will give rise to a performance guarantee as a function of  $F(\{\emptyset\}_{r \in R})$, the safety of the initial allocation without any tasks assigned but with only the goal of reaching the exit. This could potentially take a large safety value as no task is needed to be done, e.g., 1, which could deteriorate the performance guarantee. To address this issue, we introduce the reverse greedy algorithm next.

\vspace{-.1cm}
\subsection{Reverse greedy algorithm}\label{sec:reverse}
The reverse greedy algorithm is initialized with all tasks being allocated to every robot simultaneously. Due to this reason, its performance guarantee will instead be a function of $F\left(\{T\}_{r \in R}\right)$. This algorithm iteratively updates its provisional allocation by removing tasks from the robots.
In each step, the task-robot pair causing the largest optimality loss is removed. It converges when every task is allocated to exactly one robot. As is the case for the forward greedy, computation of the iteration steps is distributed among the robots. \revi{Since the reverse greedy implementation uses similar principles to those found in the forward, its detailed description is relegated to Appendix~\ref{sec:apx_reverse_greedy}.}

\subsection{Performance guarantees}\label{sec:guarantees}
To discuss the theoretical performance guarantees for Algorithms~\ref{alg:fg_distributed} and~\ref{alg:rg_distributed}, we bring in the definitions for the curvature $\alpha$ and the submodularity ratio $\gamma$.

\begin{definition}\label{def:curvature}
\textit{Curvature} of a nonincreasing $F$ is the smallest $\alpha \in \R_+$ such that
$(1-\alpha) \cdot \left[ F(B \cup \{e\})-F(B)\right] \geq F(A \cup \{e\})-F(A),$
for all $A \subseteq B \subseteq W$, for all $e \in W \setminus B$. Observe that $F$ is supermodular if and only if $\alpha=0$, and we also have $\alpha\in [0,1]$. Refer to \cite{bian2017guarantees} for derivations.
\end{definition}

\begin{definition}\label{def:submodularityratio}
\textit{Submodularity ratio} of a nonincreasing $F$ is the largest $\gamma \in \R_+$ such that $ \gamma \cdot \left[F(A \cup \{e\})-F(A)\right]\geq F(B \cup \{e\})-F(B)$
for all $A \subseteq B \subseteq W$, for all $e \in W \setminus B$. Observe that $F$ is submodular if and only if $\gamma=1$, and we also have $\gamma \in [0,1]$. Refer to \cite{bian2017guarantees} for derivations.
\end{definition}

\revi{Many set function optimization problems take advantage of submodularity or supermodularity properties of their objective functions. However, we will see in the numerics that our objective function does not exhibit these properties. Instead, we bring in the submodularity ratio and curvature properties above describing how far a nonsubmodular or nonsupermodular set function is from being submodular or supermodular, respectively. These ratios are used to define weak notions of these properties.} Calculating these values is computationally expensive. Moreover, unlike many past studies on these notions~\cite{bian2017guarantees}, the multiplicative group safety we consider is not amenable to ex-ante bounds on these ratios. \revi{In our numerical case studies, we will verify our objective function $F$ to be strictly decreasing after each additional task assignment.\footnote{This holds unless the new task appears to be already on an optimal path.} It is known that strict monotonicity is a sufficient condition for non-trivial values for submodularity ratio ($\gamma>0$) and curvature ($\alpha<1$)~\cite{guo2020actuator}. Moreover, similar to~\cite{guo2020actuator}, we will compute ex-post bounds from the function evaluations obtained from the greedy algorithms. Obtaining $\gamma<1$ and $\alpha>0$ in our case studies will prove that the objective function is both nonsubmodular and nonsupermodular.}

Let $F^*$ denote the optimal value of~\eqref{eq:task_allocation_problem}. \revi{We can now invoke two performance guarantees from our recent work.}
\begin{theorem}\label{thm:guar_fg}\cite[Thm. 1]{orcun2021performance}
    Let $F^\text{fg}$ denote the objective of the forward greedy solution from Algorithm~\ref{alg:fg_distributed}. We then have
    \begin{align*}
        \frac{F^\text{fg}-F\left(\{\emptyset\}_{r \in R}\right)}{F^*-F\left(\{\emptyset\}_{r \in R}\right)}\leq \frac{1}{\gamma \cdot (1-\alpha)}.
    \end{align*}
\end{theorem}
\vspace{.1cm}

\begin{theorem}\label{thm:guar_rg}\cite[Thm. 2]{orcun2021performance}
    Let $F^\text{rg}$ denote the objective of the reverse greedy solution from Algorithm~\ref{alg:rg_distributed}. We then have
    \begin{align*}
        \frac{\gamma}{1+\gamma \cdot \alpha}\leq\frac{F^\text{rg}-F\left(\{T\}_{r \in R}\right)}{F^*-F\left(\{T\}_{r \in R}\right)}.
    \end{align*}
\end{theorem}
\vspace{.1cm}

\textbf{Comparison of the two performance guarantees.}  Both guarantees involve the objective evaluated at their initial step as a reference. For the forward greedy, this is the empty allocation, which entails that the robots are safe with high probability. In other words, safety objective $F\left(\{\emptyset\}_{r \in R}\right)\approx 1$ generally takes a high value. For the reverse greedy, this is the fully redundant allocation resulting in a low probability of success $F\left(\{T\}_{r \in R}\right)\approx 0$, since the robots are overwhelmed with  the tasks and with the increased danger of being contaminated. Assuming these values, the guarantees are 
\begin{equation}
\begin{split}
     &\frac{F^*}{\gamma \cdot (1-\alpha)}+\frac{\gamma \cdot (1-\alpha)-1}{\gamma \cdot (1-\alpha)}=g^\text{fg}(\alpha,\gamma,F^*)\leq F^\text{fg},\label{eq:rearranged_guarantees}\\
    &F^* \cdot \frac{\gamma}{1+\gamma \cdot \alpha}=g^\text{rg}(\alpha,\gamma,F^*)\leq F^\text{rg}.
 \end{split}   
\end{equation}
In~\eqref{eq:rearranged_guarantees},  $g^\text{fg}(\alpha,\gamma,F^*)$ and $g^\text{rg}(\alpha,\gamma,F^*)$ provide lower bounds on the probabilities of success: $F^\text{fg}$ and $F^\text{rg}$. We can already see that the reverse greedy guarantee directly relates to the optimal value $F^*$, where as the forward greedy guarantee can even take negative values due to additional terms. This will imply better reverse greedy guarantees in many cases.

\begin{figure}[t!]
    \centering\vspace{.1cm}
    \includegraphics[width=0.26\linewidth]{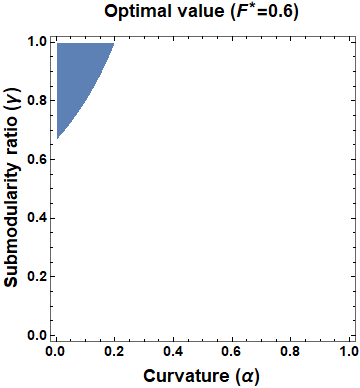}\hspace{.5cm} 
    \includegraphics[width=0.26\linewidth]{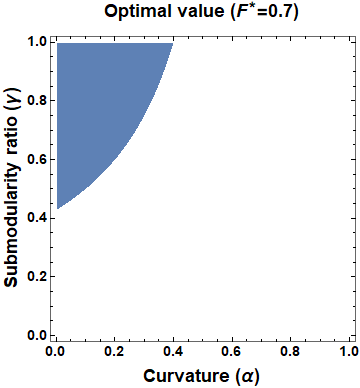}
    
    \vspace{.1cm}
    
    \includegraphics[width=0.26\linewidth]{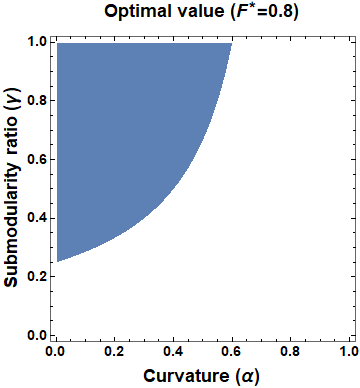}\hspace{.5cm}
    \includegraphics[width=0.26\linewidth]{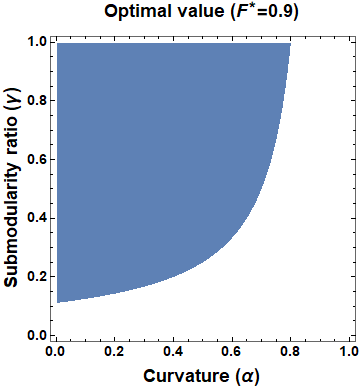}
    \caption{Comparison of guarantees. For each optimal $F^*$ value, the shaded regions represent the $(\alpha, \gamma)$ pairs for which the forward greedy outperforms the reverse greedy.}
    \label{fig:table}\vspace{-.2cm}
\end{figure}

 Figure~\ref{fig:table} illustrates the area defined by $\{(\alpha,\gamma) \in [0,1]\times[0,1]\,\vert\,g^\text{fg}(\alpha,\gamma,F^*)\geq g^\text{rg}(\alpha,\gamma,F^*)\}$ for fixed values of $F^*$. Observe that if $F^*$ is close to $1$, using the forward greedy is a better choice for a range of values of $\alpha$ and $\gamma$. However, if $F^*\leq 0.5$, the reverse greedy provides a better guarantee for all possible $(\alpha, \gamma)$ pairs. As the value of $F^*$ decreases from $1$ to $0.5$,  the area where the forward greedy algorithm is more reliable shrinks. For these range of values, the performance guarantee of the forward greedy algorithm is better only when the function is close to being both supermodular and submodular. In fact, \cite[Prop. 4 and 5]{guo2020actuator} prove that there is no performance guarantee for the forward greedy algorithm unless both the submodularity ratio and the curvature are utilized. Hence, in this case, if one of the two properties does not hold, we expect the reverse greedy algorithm to perform better. To conclude, unless $F^*$ is expected to be sufficiently large, theory suggests implementing the reverse greedy algorithm for a larger range of problem instances. Computational comparison is provided with the case studies.

\vspace{-.15cm}
\section{Numerical results}\label{sec:case_studies}\vspace{-.15cm}
We present a case study for the two-stage multi-robot safe planning framework in Section~\ref{sec:framework}. The code is available at {\href{http://github.com/TihanyiD/multi_alloc}{\textcolor{blue}{github.com/TihanyiD/multi\_alloc}}}.
For the high-level task allocation stage, we implement the forward and the reverse greedy from Section~\ref{sec:greedy}. The example is tailored such that we can compute the optimal allocation via brute force for performance comparison. This brute force solution to \eqref{eq:task_allocation_problem} is obtained by enumerating all possible task allocations. For larger examples, the optimal allocation cannot be computed.\looseness=-1

\begin{figure}[t!] 
    \centering\vspace{.1cm}
    \includegraphics[width=0.62\linewidth]{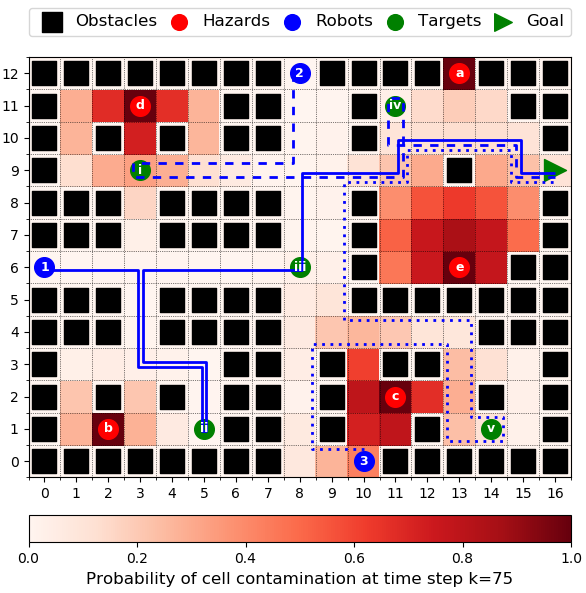} 
    \caption{Example environment and generated robot paths for the brute force optimal task allocation.}\label{fig:map_1_paths}\vspace{-.1cm} 
\end{figure}

The environment is a $17$-by-$13$ grid map with the initial state in Figure~\ref{fig:map_1_paths}. The time horizon is $N=75$ steps, which is long enough for each robot to visit all the targets in the grid. The robot dynamics are defined to be deterministic. Five hazard sources are illustrated by their initial positions in Figure~\ref{fig:map_1_paths}. The hazard dynamics are described in \revi{Appendix~\ref{sec:apx_tau_Y_model} and in \cite{wood2016automaton}} and are based on fire propagation models \cite{beer1990fire,tymstra2010development}. To visualize the evolution of the hazard, the heat map in Figure~\ref{fig:map_1_paths} shows the probability of a grid cell being \revi{hazardous} within 75 steps.\looseness=-1
\begin{table}[t!] 
    \centering\vspace{-.1cm}
    \caption{Comparison of task allocation algorithms. \textit{Task allocation} -- Allocation computed by the given algorithm, \textit{Computation time} -- Total algorithm run time\protect\footnotemark, \textit{Success probability} -- Corresponds to the safety objective of \eqref{eq:task_allocation_problem}}\label{tab:results_1}
   \resizebox{.35\textwidth}{!}{ \begin{tabular}{| c || c c | c | c |}
        \hline
        Algorithm & \multicolumn{2}{ c |}{\shortstack{Task\\ allocation}} & \shortstack{Computation\\ time} & \shortstack{Success\\ probability}\\
        \hline\hline
        \multirow{4}{*}{\shortstack{Forward\\ Greedy}} & Robot & Tasks & \multirow{4}{*}{\shortstack{7 minutes\\5 seconds}} & \multirow{4}{*}{0.699}\\
        \cline{2-3}
        & $1$ & $\{\text{i},\text{ii},\text{iii}\}$ & & \\
        & $2$ & $\{\text{iv}\}$ & & \\
        & $3$ & $\{\text{v}\}$ & & \\
        \hline
        \multirow{4}{*}{\shortstack{Reverse\\ Greedy}} & Robot & Tasks & \multirow{4}{*}{\shortstack{29 minutes}} & \multirow{4}{*}{0.717}\\
        \cline{2-3}
        & $1$ & $\{\text{ii},\text{iii}\}$ & & \\
        & $2$ & $\{\text{i},\text{iv}\}$ & & \\
        & $3$ & $\{\text{v}\}$ & & \\
        \hline
        \multirow{4}{*}{\shortstack{Brute\\ Force}} & Robot & Tasks & \multirow{4}{*}{\shortstack{4 hours\\53 minutes\\19 seconds}} & \multirow{4}{*}{0.717} \\
        \cline{2-3}
        & $1$ & $\{\text{ii},\text{iii}\}$ & & \\
        & $2$ & $\{\text{i},\text{iv}\}$ & & \\
        & $3$ & $\{\text{v}\}$ & & \\
        \hline
    \end{tabular}}
\end{table}
\footnotetext{Measured on a computer equipped with Core i7 (2.6GHz), 8GB RAM.}

\textbf{Performance results.} Table~\ref{tab:results_1} compares the solutions from three different task allocation methods. The optimal task allocation and the corresponding robot paths can be found in Figure~\ref{fig:map_1_paths}. The forward and the reverse greedy algorithms provide tractable approximations of the optimal solution. The computation times illustrate the clear benefits of greedy. Both greedy algorithms are faster than the brute force by order of magnitude without significant optimality loss (in reverse greedy, there is no loss at all). 

The reverse greedy algorithm takes significantly more time than the forward greedy. We explain this by the following points. First, in Algorithm~\ref{alg:fg_distributed}, the forward greedy takes $|T|$ steps, whereas the reverse greedy \revi{(Algorithm~\ref{alg:rg_distributed})} takes $|T| \cdot (|R|-1)$ steps. Second, the computation time for evaluating the solution of the single-robot safe planning problem (see Section~\ref{sec:low_level}) for a subset of tasks $T_r \subset T$ depends greatly on the number of targets $|T_r|$ allocated to a robot. The target execution state $|Q|=2^{|T_r|}$ grows exponentially with $|T_r|$, which has a major effect on computational complexity. The forward greedy explores the cases where a smaller number of tasks are allocated to the robots, as it is initialized with no tasks assigned to any robot. In contrast, the reverse greedy is initialized with all tasks being allocated to every robot and removes tasks gradually. Hence, the reverse greedy requires solving larger instances of the single-robot safe planning problem. \looseness=-1 

Notice that planning for the shortest path might jeopardize safety. For example, for `Robot 3', the shortest path between its initial cell and `target v' would  pass through `hazard c'. Thus, a safe planning framework, such as the one proposed here, is crucial for accounting for dynamic uncertainties.

\textbf{Properties of the safety objective and the guarantees.} 
There is no computationally tractable approach to obtaining the exact curvature $\alpha$ and the exact submodularity ratio $\gamma$ properties of $F$ (see Definitions~\ref{def:curvature} and~\ref{def:submodularityratio}). However, we can confirm that these ratios are non-trivial, $\alpha<1$ and $\gamma>0$, because we numerically verified that $F$ is strictly decreasing for our case study. Moreover, we obtained ex-post bounds called \textit{greedy-approximate curvature} $\alpha^G \leq \alpha$ and \textit{greedy-approximate submodularity ratio} $\gamma^G \geq \gamma$, see their definitions in \cite{bian2017guarantees}. We calculated these bounds using only the function evaluations during the execution of the greedy algorithms instead of taking all possible task allocations into account. In literature, they are commonly used as computationally efficient alternatives to these ratios~\cite{bian2017guarantees}. For this particular example, we obtained the values $\alpha^G=0.989$ and $\gamma^G=0.525$. \revi{Since $\alpha^G>0$ and $\gamma^G<1$, we can verify that the objective function is indeed nonsupermodular and nonsubmodular, respectively (see Definitions~\ref{def:curvature} and~\ref{def:submodularityratio}).} Although
the guarantees of Theorems~\ref{thm:guar_fg} and~\ref{thm:guar_rg} do not necessarily hold for $\alpha^G$ and $\gamma^G$, evaluating \eqref{eq:rearranged_guarantees} at $\alpha^G$ and $\gamma^G$ suggests that the reverse greedy may essentially have a better performance guarantee than the forward greedy for this particular problem instance. This can be attributed to the fact that the function~$F$ is far away from being supermodular, $\alpha^G=0.989$, and this heavily deteriorates the bound in Theorem~\ref{thm:guar_fg}. 
This observation is confirmed by the empirical performances in Table~\ref{tab:results_1}.

\textbf{Additional results.}
We provide two studies in appendices. The first involves an example where the forward greedy can perform better than the reverse greedy, even though the guarantees suggest otherwise.
As a remark, both algorithms are close to optimal in this particular instance.
The second is a large set of randomized examples for different number of robots and tasks. For the same robots and tasks, the forward converges faster than the reverse in all cases. In the case of the forward, the computation times can decrease as the number of robots increase as this implies avoiding solving the path planning for a high number of allocated tasks.


\section{Conclusion}\label{sec:conclusion}
We proposed a two-stage framework to solve a multi-robot safe planning problem in a tractable manner. An efficient implementation of a stochastic reachability for a Markov decision process addressing safe planning under dynamic uncertainties served as the low-level planner. {The multiplicative safety objective allowed implementations of the forward and reverse greedy in a distributed auction-based manner to allocate the tasks among the robots.} Through case studies, we compared our solutions with the computationally intractable optimal solution. We illustrated that our algorithms perform well both in terms of computation time and optimality. The reverse greedy can have a better performance guarantee than the forward greedy, but this benefit came with an increased computational burden. Future works include accounting for robot failure, and incorporating hazard observation feedback.\looseness=-1

\setstretch{.9}
\bibliographystyle{IEEEtran}
\bibliography{bibliography}
\setstretch{1}

\newpage
\appendix

\subsection{Neighborhood-based hazard model}\label{sec:apx_tau_Y_model}
In the following, we propose the neighborhood based hazard dynamics $\tau_Y$ based on \cite[\S IV./C.]{wood2016automaton}. Let $y^k \in Y$ be the hazard state and $x \in X \setminus y^k$ be a nonhazardous cell at time step $k$. We also introduce the scalar parameter $\theta \in [0,1]$ as the \textit{spread speed} parameter which controls the speed of the evolving hazard. Consider that $x$ can be ignited by any of its hazardous direct neighbours $\bar{x}_N \in N(x) \cap y^k$ with probability $\theta$ and hazardous diagonal neighbours $\bar{x}_D \in D(x) \cap y^k$ with probability $\theta \mathbin{/} \sqrt{2}$, to take the distance into account. Let $n_N(x,y^k)=|N(x) \cap y^k|$ and $n_D(x,y^k)=|D(x) \cap y^k|$ denote the numbers of direct and diagonally hazardous neighbours of $x$. Now, we can define the following function $p_{nc}: X \times Y \rightarrow [0,1]$ as the probability of nonhazardous position $x$ remaining nonhazardous given hazard state $y^k$ at step $k$:
\begin{equation*}
    p_{nc}(x\,\vert\,y^k)=\left(1-\theta\right)^{n_N(x,y^k)} \cdot \left(1-\frac{\theta}{\sqrt{2}}\right)^{n_D(x,y^k)}.
\end{equation*}
Note that the smaller the value of $\theta$, the less likely it is for position $x$ becomes hazardous, hence the slower the hazard spreads. Once the hazard reaches a cell, it remains hazardous throughout the whole process. In order to model this behaviour, we define $p_c: X \times Y \rightarrow [0,1]$ representing the probability of position $x$ getting hazardous given hazard state $y^k$ at step $k$ the following way
\begin{equation*}
    p_c(x\,\vert\,y^k)=
    \begin{cases}
        1-p_{nc}(x\,\vert\,y^k) &\text{if} \hsp x \notin y^k,\\
        1 &\text{if} \hsp x \in y^k.
    \end{cases}
\end{equation*}
Based on the above, we  define the transition kernel $\tau_Y: Y \times Y \rightarrow [0,1]$ as
\begin{equation*}
    \tau_Y\left(y^{k+1}\,\vert\,y^k\right)=\prod\limits_{x \in y^{k+1}} p_c(x\,\vert\,y^k)\cdot \prod\limits_{x \in (X\setminus y^{k+1})} 1-p_c(x\,\vert\,y^k).
\end{equation*}
This is a well-defined kernel, as it sums to one over all $y^{k+1}$, and it is non-negative.

\subsection{Description of transition probabilities}\label{sec:apx_tau_S_k}
\textit{Robot motion} -- Defined by $\tau_X(x^{k+1}\,\vert\,x^k,u^k)$ in Sec.~\ref{sec:modelling}.

\textit{Task execution state transition} -- The transition at time step $k$ from $q^k \subseteq Q$ to $q^{k+1}\subseteq Q$ given that the robot is at position $x^{k+1}$ at step $k+1$: This is described by the following time-homogeneous transition kernel
    \begin{equation*}
        \tau_Q(q^{k+1}\,\vert\,q^k,x^{k+1})=
        \begin{cases}
            1 & \text{if} \hsp q^{k+1}=q^k \cup (x^{k+1} \cap T_r),\\
            0 & \text{otherwise},
        \end{cases}
    \end{equation*}
    where $\tau_Q: Q \times Q \times X \rightarrow [0,1]$. The interpretation of this transition kernel is as follows. Whenever a target position $x^{k+1} \in T_r$ is visited, it is added to the list $q^{k}$ to form $q^{k+1}$. For any non-target position $x^{k+1} \notin T_r$, we have $x^{k+1} \cap T_r=\emptyset$ and $q^{k+1}=q^k$. If a target position $x^{k+1} \in T_r$ is visited more than once, since $x^{k+1} \in q^k$, we have $q^{k+1}=q^k$.

\textit{Contamination probability} -- As discussed, since the hazard state $y^k$ can potentially be any subset of $X$, the computation of its transition kernel $\tau_Y$ is intractable. Hence, the control policy of the robot cannot depend explicitly on the hazard state. However, the robot can account for the initial hazard state $y^0$ and its evolution dynamics $\tau_Y$ to evaluate the probability of a given trajectory being contaminated during the planning phase. To this end, we define the function $p_H^k: X \times X \rightarrow [0,1]$,  such that the value of $p_H^k(x^{k+1},x^k)$ equals to the  probability of transitioning from a \revi{nonhazardous cell} $x^k \notin y^k$ at time $k$ to a \revi{hazardous cell} $x^{k+1} \in y^{k+1}$ at time $k+1$. We use Monte-Carlo simulation to evaluate this function during planning, given $y_0$ and $\tau_Y$.\footnote{The precise evaluation of the function $p_H^k$ is computationally intractable due to the size of $Y=2^X$. In \cite[Alg. 1]{lu2020safe}, a Monte-Carlo sampling based algorithm was proposed to provide a tractable approximation of $p_H^k$. The idea is to forward propagate the hazard dynamics using $y_0$ and kernel $\tau_Y$. }

With the elements above, we can now define the transition kernel $\tau_S^k$ at time step $k$ as follows
\revi{\medmuskip=1.2mu\thinmuskip=1.2mu\thickmuskip=1.2mu\begin{equation*}\label{eq:tau_S_k}
\begin{split}
    &\tau_S^k(s^{k+1}\,\vert\,s^k,u^k)\,=\,\vspace{.1cm}\\\vspace{.1cm}
    &\begin{cases}
        1 \hspace{3.5cm} \text{if} \hsp s^{k+1}=s^k, s^k\in \{s_G,s_H\},\vspace{2mm}\vspace{.68cm}\\
      \sum\limits_{x^{k+1} \in X} p_H^k(x^{k+1},x^k)
        \cdot\tau_X(x^{k+1}\,\vert\,x^k,u^k)\\ \hspace{3.7cm} \text{if} \ (s^{k+1}=s_H) \wedge s^k \notin \{s_G,s_H\},\vspace{2mm}\vspace{.68cm}\\
        \hspace{.1cm} \left(1-p_H^k(x^{k+1},x^k)\right)         \cdot \tau_Q(q^{k+1}\,\vert\,q^k,x^{k+1}) \cdot \tau_X(x^{k+1}\,\vert\,x^k,u^k) \vspace{.1cm}\\ \hspace{3.7cm} \text{if} \hsp (s^{k+1}\neq s_H) \wedge s^k \notin \{s_G,s_H\},\vspace{2mm}\vspace{.68cm}\\
        0 \hspace{3.5cm} \text{otherwise.}
    \end{cases}
    \end{split}
\end{equation*}}
The transition kernel captures that the goal state $s_G$ and the hazard state $s_H$ are  \textit{absorbing}. Once they are reached, the system state does not change anymore due to either contamination or mission completion. In any other state $s^k\notin \{s_G,s_H\}$, the robot can move to a different state following the dynamics defined by the transition kernel above.

\subsection{Reverse greedy algorithm}\label{sec:apx_reverse_greedy}

In this appendix, we provide the description of the reverse greedy algorithm. Algorithm~\ref{alg:rg_distributed} proceeds as follows. First, define the following variables for each step $k$: $\{T_r^k\}_{r \in R}$ denotes the current task allocation, while $\{f_r^k\}_{r \in R}$ stores the evaluated function values for each robot $r$. Furthermore, $J^k$ is the set of tasks not yet removed and $R^k$ is the set of robots which need to update their bids in the next step. We initially assign all tasks to every robot, hence $T_r^0=T$ for all $r \in R$ (see Line~\ref{alg:rg_distributed_line_1}). In each step, exactly one task is removed from one of the robots. Hence, the algorithm needs $|T| \cdot (|R|-1)$ steps to complete (Line~\ref{alg:rg_distributed_line_2}). At each $k$, the robot $r \in R$ submits a bid (see Line~\ref{alg:rg_distributed_line_3}--\ref{alg:rg_distributed_line_5}), which consists of the pair $(t_r^k,\delta_r^k)$. Each robot $r$ submitting a bid chooses the task $t_r^k$ from the not yet removed tasks $J^{k-1} \cap T_r^{k-1}$ that corresponds to the largest optimality loss $\delta_r^k$ with respect to the individual objective function $f_r$. After collecting all the bids, we choose the robot $r^k$ which generates the largest optimality loss with respect to the collective objective: the multiplicative group safety $F$ (Line~\ref{alg:rg_distributed_line_6}). Due to our auction-based formulation in this line, we can choose the task-robot pair with the largest collective loss efficiently. Between Lines~\ref{alg:rg_distributed_line_7}--\ref{alg:rg_distributed_line_8}, we set the values of $f_r^k$, $T_r^k$ for all $r \in R$ and $R^k$, $J^k$ according to our choice of task allocation. Note that a task $t_{r^k}^k$ is removed from $J^k$ if it is only allocated to single robot, hence $|\{r|t_{r^k}^k \in T_r^k\}|=1$. Observe that only the robots choosing the same task as $r^k$ (i.e. $r$ such that $t_r^k=t_{r^k}^k$) have to update their bids in the next iteration and this happens only right after $t_{r^k}^k$ gets removed from $J^k$ (see Line~\ref{alg:rg_distributed_line_9} at step $k$ and Line~\ref{alg:rg_distributed_line_3} at step $k+1$). The rest of the robots simply resubmit their bids from the previous iteration (see Line~\ref{alg:rg_distributed_line_5}). The variable $R^k$ is initialized with $R^0=R$, since in the first iteration all robots have to calculate their bids.

\begin{algorithm}[t]
    \setstretch{.95}
    \setcounter{AlgoLine}{0}
	\KwIn{$R$, $T$, $\{f_r\}_{r \in R}$} \KwOut {$\{T_r^\text{rg}=T_r^{|T| \cdot (|R|-1)}\}_{r \in R}$}
	\SetKwBlock{Begin}{begin}{end}
	\Begin{
		initialization: \mbox{$T_r^0=T$, $f_r^0=f_r(T)$, $\forall r$, $J^0=T$, $R^0=R$}\label{alg:rg_distributed_line_1}\\
		\For{$k=1,\dots,|T| \cdot (|R|-1)$}{ \label{alg:rg_distributed_line_2}
			\For{$r \in R^{k-1}$}{\label{alg:rg_distributed_line_3}
			    \mbox{$t_r^k \gets \argmax\limits_{t \in J^{k-1} \cap T_r^{k-1}} f_r(T_r^{k-1} \setminus t)-f_r(T_r^{k-1})$}\\
			    
				\mbox{$\delta_r^k \gets f_r(T_r^{k-1} \setminus t_r^k)-f_r(T_r^{k-1})$}\label{alg:rg_distributed_line_4}
			}
			\mbox{$(t_r^k,\delta_r^k) \gets (t_r^{k-1},\delta_r^{k-1}) \hspace{2mm} \forall r \notin R^{k-1}$}\label{alg:rg_distributed_line_5}\\
			
			\mbox{$r^k \gets \argmax_{r \in R} \delta_r^k \cdot \prod_{r' \in R\setminus\{r\}} f_{r'}^{k-1}$}\label{alg:rg_distributed_line_6}\\
			
			\mbox{$f_r^k \gets 
		f_r^k+\delta_r^k \text{, if } r=r^k;\
					f_r^{k-1} \text{, otherwise}$}\label{alg:rg_distributed_line_7}\\
				
			\mbox{$	T_r^k \gets 
    				T_r^k \setminus t_r^k \text{, if } r=r^k;\
    				T_r^{k-1} \text{, otherwise}$}\\
				
			\mbox{$R^k \gets
				\begin{cases}
					\left\{r \,\rvert\, t_r^k=t_{r^k}^k\right\} \text{, if } \bigl | \left\{r \,\rvert\, t_{r^k}^k \in T_r^k\right\} \bigr |=1\\
					\{r^k\} \text{, otherwise}
				\end{cases}$}\label{alg:rg_distributed_line_9}\\
				
			\mbox{$J^k \gets
				\begin{cases}
					J^{k-1} \setminus t_{r^k}^k \text{, if } \bigl | \left\{r \,\rvert\, t_{r^k}^k \in T_r^k\right\} \bigr |=1\\
					J^{k-1} \text{, otherwise}
				\end{cases}$}\label{alg:rg_distributed_line_8}
		}
	}
	\caption{Reverse Distributed Greedy Algorithm}\label{alg:rg_distributed}
\end{algorithm}

\subsection{Additional numerical case studies}\label{sec:apx_case_studies}

We present two additional numerical example pairs using the hazard evolution models and robot motion dynamics described in Section~\ref{sec:case_studies}. The parameters of these examples are designed in the following way to illustrate the empirical performance and the theoretical guarantee comparisons for the forward and the reverse greedy algorithms.

Examples 2.1 and 2.2 show the same map where in the case of Example 2.1 the hazard spreads with a higher probability. Hence, for our comparisons, in Example 2.1, the environment can be said to be more challenging for the robots. The same distinction applies also to the pair of Examples 3.1 and 3.2. In all four examples the reverse greedy enjoys a better theoretical guarantee than the forward greedy algorithm. In Examples 2.1 and 2.2, the reverse greedy indeed outperforms the forward in terms of empirical performance. However, in the case of Examples 3.1 and 3.2, the forward greedy achieves a better empirical performance. This is intended to show that theoretical guarantees define lower bounds on the worst-case algorithm performances. Even if an algorithm enjoys a better theoretical guarantee, it does not necessarily achieve a better empirical performance.

\begin{figure}[t!]
    \centering\vspace{.1cm}
    \includegraphics[width=0.9\linewidth]{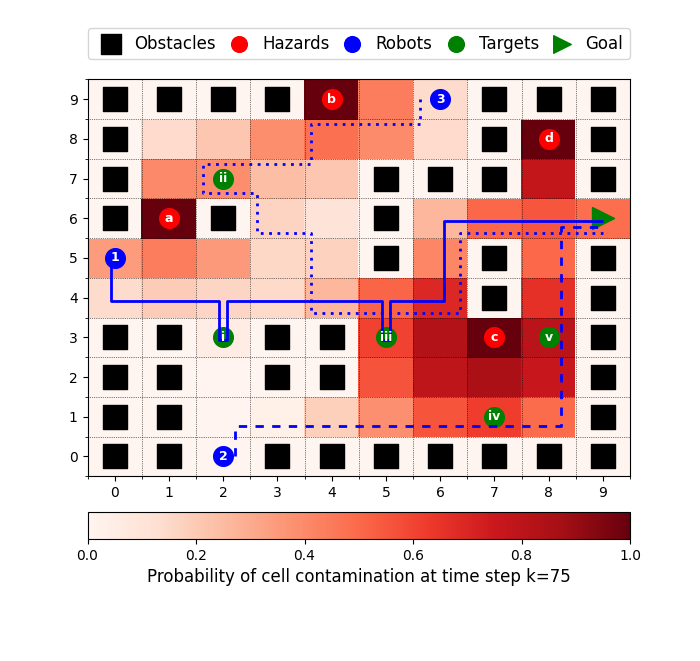}\hspace{.5cm} 
    \includegraphics[width=0.9\linewidth]{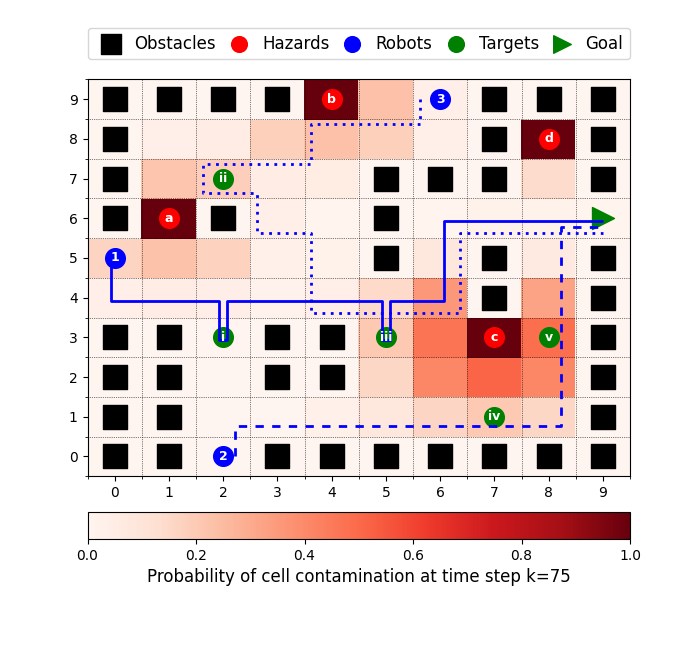}
    \caption{Example 2.1 and 2.2 setup, respectively. The plots also show the brute force optimal task allocation and the corresponding optimal robot paths.}
    \label{fig:map_example_2}
\end{figure}

\textbf{Examples 2.1 and 2.2.} We implemented an example pair involving the same map, robot and target locations. The only difference between the two examples is the hazard spread parameters. In case of Example 2.1, the hazard spreads with higher probability, hence the environment is more challenging to the robots. This is visualized in Figure~\ref{fig:map_example_2}. 

 The $F^*$ value shows the multiplicative group safety in case of the brute force optimal task allocation. The examples illustrate the effect of different $F^*$ values on the outcome of the forward and reverse greedy algorithms. In case of Example 2.1 the environment is more challenging to the robots with $F^*<0.5$. While in case of Example 2.2, the environment is less challenging and $F^*>0.5$.

\begin{table}[t!] 
    \centering\vspace{-.1cm}
    \caption{Comparison of task allocation algorithms. \textit{Task allocation} -- Allocation computed by the given algorithm, \textit{Computation time} -- Total algorithm run time\protect, \textit{Success probability} -- Corresponds to the safety objective of \eqref{eq:task_allocation_problem}.}\label{tab:results_2}
   \resizebox{\linewidth}{!}{ \begin{tabular}{| c || c c | c | c || c c | c | c |}
        \hline
        \multirow{2}{*}{Algorithm} & \multicolumn{4}{ c ||}{Example 2.1} & \multicolumn{4}{ c |}{Example 2.2}\\
        \cline{2-9}
        & \multicolumn{2}{ c |}{\shortstack{Task\\ allocation}} & \shortstack{Computation\\ time} & \shortstack{Success\\ probability } & \multicolumn{2}{ c |}{\shortstack{Task\\ allocation}} & \shortstack{Computation\\ time} & \shortstack{Success\\ probability}\\
        \hline\hline
        \multirow{4}{*}{\shortstack{Forward\\ Greedy}} & Robot & Tasks & \multirow{4}{*}{\shortstack{33 seconds}} & \multirow{4}{*}{0.359} & Robot & Tasks & \multirow{4}{*}{\shortstack{46 seconds}} & \multirow{4}{*}{0.660}\\
        \cline{2-3}\cline{6-7}
        & $1$ & $\{\text{iii}\}$ & & & $1$ & $\{\}$ & & \\
        & $2$ & $\{\text{i},\text{iv},\text{v}\}$ & & & $2$ & $\{\text{i},\text{iii},\text{iv},\text{v}\}$ & & \\
        & $3$ & $\{\text{ii}\}$ & & & $3$ & $\{\text{ii}\}$ & & \\
        \hline
        \multirow{4}{*}{\shortstack{Reverse\\ Greedy}} & Robot & Tasks & \multirow{4}{*}{\shortstack{3 minutes\\ 25 seconds}} & \multirow{4}{*}{0.407} & Robot & Tasks & \multirow{4}{*}{\shortstack{3 minutes\\ 50 seconds}} & \multirow{4}{*}{0.719}\\
        \cline{2-3}\cline{6-7}
        & $1$ & $\{\text{i},\text{iii}\}$ & & & $1$ & $\{\text{i},\text{iii}\}$ & & \\
        & $2$ & $\{\text{iv},\text{v}\}$ & & & $2$ & $\{\text{iv},\text{v}\}$ & & \\
        & $3$ & $\{\text{ii}\}$ & & & $3$ & $\{\text{ii}\}$ & &\\
        \hline
        \multirow{4}{*}{\shortstack{Brute\\ Force}} & Robot & Tasks & \multirow{4}{*}{\shortstack{29 minutes\\48 seconds}} & \multirow{4}{*}{0.407} & Robot & Tasks & \multirow{4}{*}{\shortstack{31 minutes\\24 seconds}} & \multirow{4}{*}{0.719} \\
        \cline{2-3}\cline{6-7}
        & $1$ & $\{\text{i},\text{iii}\}$ & & & $1$ & $\{\text{i},\text{iii}\}$ & & \\
        & $2$ & $\{\text{iv},\text{v}\}$ & & & $2$ & $\{\text{iv},\text{v}\}$ & & \\
        & $3$ & $\{\text{ii}\}$ & & & $3$ & $\{\text{ii}\}$ & & \\
        \hline
    \end{tabular}}
\end{table}

We summarized the results in Table~\ref{tab:results_2}. The $F^*$ values of Example 2.1 and 2.2 are $0.407$ and $0.719$, respectively (see the Success probability for the Brute Force algorithm). Furthermore, we calculated the greedy submodularity ratio and curvature values for Example 2.1 $\alpha^G=0.412$, $\gamma^G=0.307$ and for Example 2.2 $\alpha^G=0.655$, $\gamma^G=0.575$ (see Definitions~\ref{def:curvature} and~\ref{def:submodularityratio}). We can now apply our theoretical guarantees introduced in Theorem~\ref{thm:guar_fg} and~\ref{thm:guar_rg} to the described parameters of Example 2.1 and 2.2 and compare the results. In both cases, the reverse greedy algorithm enjoys a higher theoretical guarantee $g^\text{rg}(\alpha,\gamma,F^*)>g^\text{fg}(\alpha,\gamma,F^*)$ (see Equation \eqref{eq:rearranged_guarantees}) which is indeed reflected in the results. However, this does not necessarily mean that reverse greedy should always perform better in terms of empirical performance. The comparison of theoretical guarantees only establish a worst-case lower bound on the optimal value and can serve as a reliability indicator rather then an empirical performance indicator.

In the following, we introduce Example 3.1 and 3.2, which highlight that algorithm enjoying higher theoretical guarantee might perform worse in terms of optimality.

\begin{figure}[t!]
    \centering\vspace{.1cm}
    \includegraphics[width=0.9\linewidth]{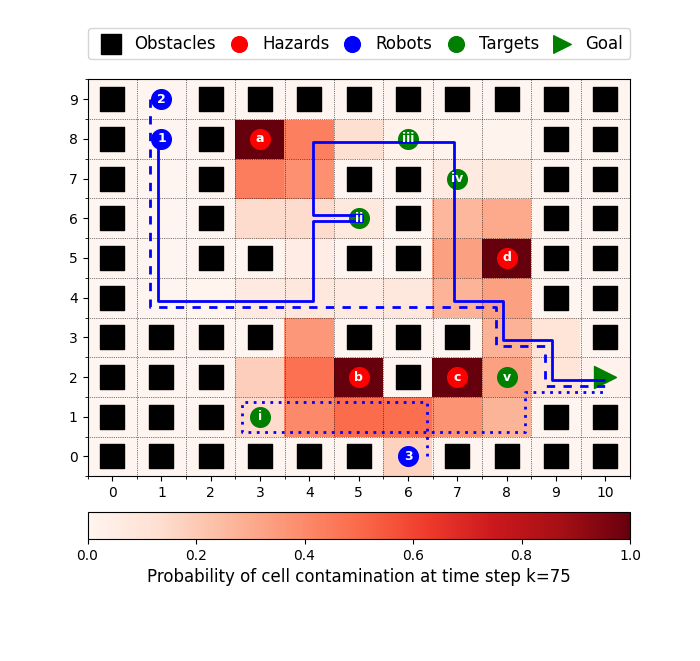}\hspace{.5cm} 
    \includegraphics[width=0.9\linewidth]{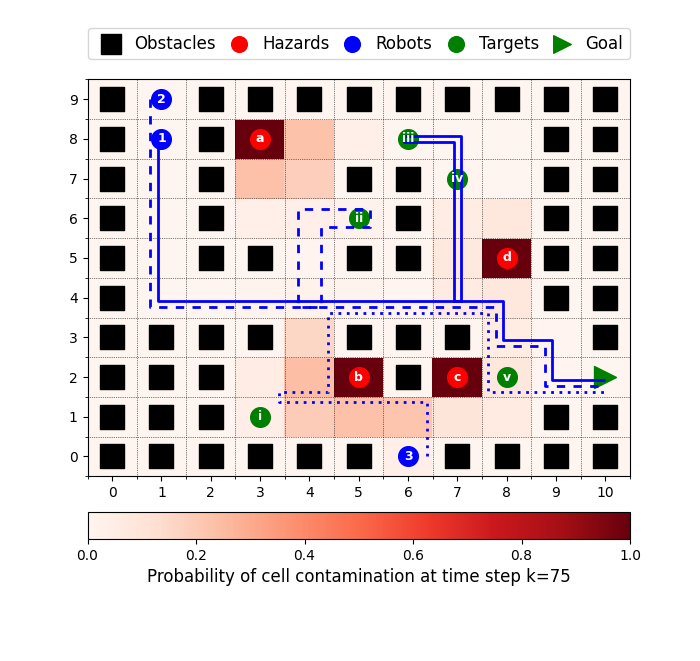}
    \caption{Example 3.1 and 3.2 setup, respectively. The plots also show the brute force optimal task allocation and the corresponding optimal robot paths.}
    \label{fig:map_example_3}
\end{figure}

\textbf{Examples 3.1 and 3.2.} Note that these examples are different from Examples 2.1 and 2.2, which are illustrated in Figure~\ref{fig:map_example_3}. Again, both examples in the pair involve the same map, robot and target locations. The only difference between the two examples is the hazard spread parameters.

\begin{table}[t!] 
    \centering\vspace{-.1cm}
    \caption{Comparison of task allocation algorithms. \textit{Task allocation} -- Allocation computed by the given algorithm, \textit{Computation time} -- Total algorithm run time\protect, \textit{Success probability} -- Corresponds to the safety objective of \eqref{eq:task_allocation_problem}.}\label{tab:results_3}
   \resizebox{\linewidth}{!}{ \begin{tabular}{| c || c c | c | c || c c | c | c |}
        \hline
        \multirow{2}{*}{Algorithm} & \multicolumn{4}{ c ||}{Example 3.1} & \multicolumn{4}{ c |}{Example 3.2}\\
        \cline{2-9}
        & \multicolumn{2}{ c |}{\shortstack{Task\\ allocation}} & \shortstack{Computation\\ time} & \shortstack{Success\\ probability } & \multicolumn{2}{ c |}{\shortstack{Task\\ allocation}} & \shortstack{Computation\\ time} & \shortstack{Success\\ probability}\\
        \hline\hline
        \multirow{4}{*}{\shortstack{Forward\\ Greedy}} & Robot & Tasks & \multirow{4}{*}{\shortstack{29 seconds}} & \multirow{4}{*}{0.364} & Robot & Tasks & \multirow{4}{*}{\shortstack{25 seconds}} & \multirow{4}{*}{0.752}\\
        \cline{2-3}\cline{6-7}
        & $1$ & $\{\text{ii}\}$ & & & $1$ & $\{\text{ii}\}$ & & \\
        & $2$ & $\{\text{iii},\text{iv}\}$ & & & $2$ & $\{\text{iii},\text{iv}\}$ & & \\
        & $3$ & $\{\text{i},\text{v}\}$ & & & $3$ & $\{\text{i},\text{v}\}$ & & \\
        \hline
        \multirow{4}{*}{\shortstack{Reverse\\ Greedy}} & Robot & Tasks & \multirow{4}{*}{\shortstack{2 minutes\\ 39 seconds}} & \multirow{4}{*}{0.354} & Robot & Tasks & \multirow{4}{*}{\shortstack{2 minutes\\ 33 seconds}} & \multirow{4}{*}{0.733}\\
        \cline{2-3}\cline{6-7}
        & $1$ & $\{\text{v}\}$ & & & $1$ & $\{\text{i},\text{iii},\text{iv}\}$ & & \\
        & $2$ & $\{\text{ii},\text{iii},\text{iv}\}$ & & & $2$ & $\{\}$ & & \\
        & $3$ & $\{\text{i}\}$ & & & $3$ & $\{\text{i},\text{v}\}$ & &\\
        \hline
        \multirow{4}{*}{\shortstack{Brute\\ Force}} & Robot & Tasks & \multirow{4}{*}{\shortstack{22 minutes\\47 seconds}} & \multirow{4}{*}{0.379} & Robot & Tasks & \multirow{4}{*}{\shortstack{21 minutes\\12 seconds}} & \multirow{4}{*}{0.753} \\
        \cline{2-3}\cline{6-7}
        & $1$ & $\{\text{ii},\text{iii},\text{iv}\}$ & & & $1$ & $\{\text{iii},\text{iv}\}$ & & \\
        & $2$ & $\{\}$ & & & $2$ & $\{\text{ii}\}$ & & \\
        & $3$ & $\{\text{i},\text{v}\}$ & & & $3$ & $\{\text{i},\text{v}\}$ & & \\
        \hline
    \end{tabular}}
\end{table}

We summarized the results in Table~\ref{tab:results_3}. The $F^*$ values of Example 3.1 and 3.2 are $0.379$ and $0.753$, respectively (see the Success probability for the Brute Force algorithm). Furthermore, we calculated the greedy submodularity ratio and curvature values for Example 3.1 $\alpha^G=0.035$, $\gamma^G=0.290$ and for Example 3.2 $\alpha^G=0.619$, $\gamma^G=0.477$ (see Definitions~\ref{def:curvature} and~\ref{def:submodularityratio}). We can now apply our theoretical guarantees introduced in Theorem~\ref{thm:guar_fg} and~\ref{thm:guar_rg} to the described parameters of Example 3.1 and 3.2 and compare the results. In both cases the reverse greedy performs worse in terms of empirical performance despite enjoying a higher theoretical guarantee $g^\text{rg}(\alpha,\gamma,F^*)>g^\text{fg}(\alpha,\gamma,F^*)$ (see Equation \eqref{eq:rearranged_guarantees}).

Finally, in all four examples (Examples 2.1, 2.2, 3.1, 3.2) the greedy algorithms perform better in terms of computation time by order of magnitude, and the forward greedy is faster than the reverse greedy. This further justifies our conclusions in Section~\ref{sec:case_studies}.

\subsection{Randomized example for statistical comparison}

\begin{figure}[t!] 
    \centering\vspace{.1cm}
    \includegraphics[width=0.8\linewidth]{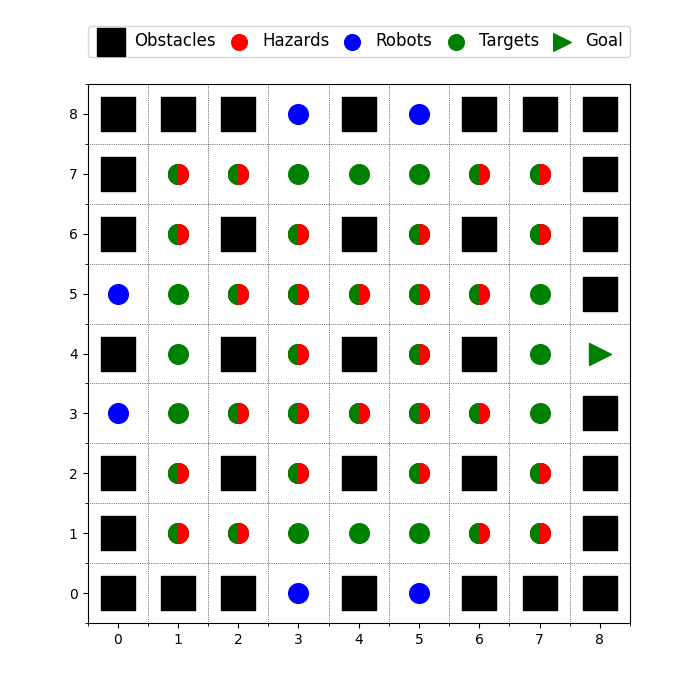} 
    \caption{Map for randomly generated examples. The colored markers (red, blue, green) correspond to the possible cells where hazards, robots or targets can randomly be placed, respectively.}\label{fig:random_map}\vspace{-.1cm}
\end{figure}

We support our findings by evaluating randomly generated examples. We use the map in Figure~\ref{fig:random_map}. We first randomly place $n_T$ targets, then $n_H$ hazards, finally $n_R$ robots onto the cells indicated by the green, red, and blue markers, respectively. We exclude cases where multiple targets, hazards or robots occupy the same cell. Furthermore, we use the neighborhood-based hazard model (Appendix~\ref{sec:apx_tau_Y_model}) with $\theta=0.02$ spread speed parameter.

We generate $20$ random examples for $n_H=3$ and each $(n_T,n_R)$ pair of the following set $\{(n_T,n_R) \in \{2,\dots,7\} \times \{2,\dots,5\} | n_R \leq n_T\}$. Note that we exclude the cases when $n_R > n_T$, since they would be equivalent with the $n_R=n_T$ case. We implement the forward and reverse greedy algorithms, and we also compute the brute force allocation.

We compare the average computation times of all three algorithms for each $(n_T,n_R)$ pair in Table~\ref{tab:stats_comp_time}. Both greedy algorithms provide a computationally more efficient solution compared to the brute force algorithm by orders of magnitude. Furthermore, the computation times of both greedy algorithms show a higher sensitivity to the number of tasks than robots. Forward greedy algorithm converges faster for the same number of robots and tasks than its reverse greedy counterpart. 

By increasing the number of tasks we can observe a significant increase in computation times. This can be explained with the problem size of the single-robot path planning algorithm (see Section~\ref{sec:low_level}). The task execution $Q=2^{T_r}$ component of the state space scales exponentially with the number of allocated tasks. The number of robots have an effect only on the number of algorithm steps. We highlight that in case of the forward greedy algorithm, the computation times can decrease as the number of robots increase. The forward greedy algorithm starts with an empty allocation for each robot $T_r=\emptyset$, and keeps adding new tasks. By increasing the number of robots, the tasks might be allocated more evenly, so that each robot executes lower number of tasks. This way we can avoid solving the single-robot path planning problem for a high number of allocated tasks. This observation does not apply to case of the reverse greedy algorithm, since it starts with a full allocation for each robot, hence the single-robot path planning problem needs to be solved for high number of allocated tasks $T_r$ regardless of the number of robots.

\begin{figure}[t!] 
    \centering\vspace{.1cm}
    \includegraphics[width=0.74\linewidth]{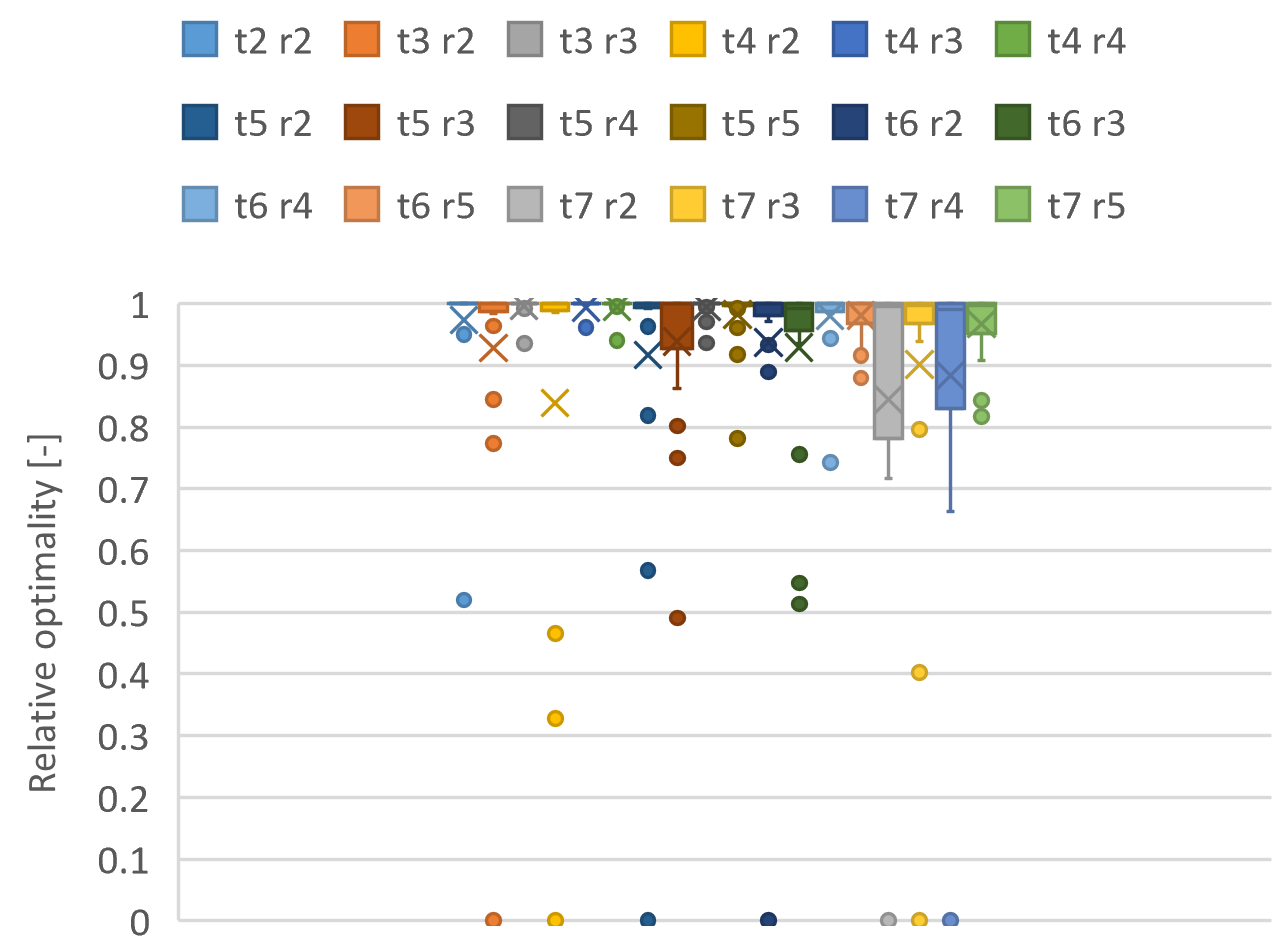} 
    \caption{Relative optimality of the best out of the two greedy algorithms (success probability of the best greedy solution relative to the brute force solution) for different number of tasks $n_T$ and robots $n_R$ (indicated by "t $n_T$ r $n_R$").}\label{fig:stats_rel_fg}\vspace{-.1cm}
\end{figure}

We show the empirical performance result of the two greedy algorithms as box plots in Figure~\ref{fig:stats_rel_fg}. We use the notion of relative optimality, the success probability of the best outcome among the two greedy algorithms divided by the brute force solution. In this figure, $\times$ denotes the average, where as $\bullet$ denotes each instance. We can conclude that greedy algorithms are not only computationally efficient but also close to being optimal in most of the larger problem instances other than the very few corner cases with low chance of success ($<5\%$) even in the brute force optimal solution.

\begin{table}[t!] 
    \centering\vspace{-.1cm}
    \caption{Average computation times of $20$ random examples for each algorithm (forward-, and reverse greedy, brute force), number of tasks $n_T$ and number of robots $n_R$.}\label{tab:stats_comp_time}
   \resizebox{\linewidth}{!}{ \begin{tabular}{| c || c | c | c  c  c  c  c  c |}
        \hline
        \multirow{6}{*}{\rotatebox[origin=c]{90}{\shortstack{Forward\\ greedy}}} & \multicolumn{2}{ c |}{} & \multicolumn{6}{ c |}{Number of tasks $n_T$}\\
        \cline{2-9}
        &  &  & 2 & 3 & 4 & 5 & 6 & 7\\ \cline{2-9}
        & \multirow{5}{*}{\rotatebox[origin=c]{90}{\shortstack{Number\\ of\\ robots\\ $n_R$}}} & 2 & $1.134$ & $2.402$ & $4.327$ & $8.802$ & $16.164$ & $48.618$\\
        \cline{3-9}
        & & 3 & * & $2.337$ & $3.983$ & $6.758$ & $12.183$ & $20.481$\\
        \cline{3-9}
        & & 4 & * & * & $4.084$ & $6.468$ & $10.773$ & $21.164$\\
        \cline{3-9}
        & & 5 & * & * & * & $7.710$ & $10.522$ & $15.327$\\
        \hline
        \hline
        \multirow{6}{*}{\rotatebox[origin=c]{90}{\shortstack{Reverse\\ greedy}}} & \multicolumn{2}{ c |}{} & \multicolumn{6}{ c |}{Number of tasks $n_T$}\\
        \cline{2-9}
        &  &  & 2 & 3 & 4 & 5 & 6 & 7\\ \cline{2-9}
        & \multirow{5}{*}{\rotatebox[origin=c]{90}{\shortstack{Number\\ of\\ robots\\ $n_R$}}} & 2 & $1.070$ & $2.309$ & $5.153$ & $15.834$ & $63.725$ & $392.223$\\
        \cline{3-9}
        & & 3 & * & $3.426$ & $7.422$ & $19.352$ & $71.724$ & $444.929$\\
        \cline{3-9}
        & & 4 & * & * & $9.625$ & $24.237$ & $79.813$ & $481.040$\\
        \cline{3-9}
        & & 5 & * & * & * & $28.875$ & $98.643$ & $602.025$\\
        \hline
        \hline
        \multirow{6}{*}{\rotatebox[origin=c]{90}{\shortstack{Brute\\ force}}} & \multicolumn{2}{ c |}{} & \multicolumn{6}{ c |}{Number of tasks $n_T$}\\
        \cline{2-9}
        &  &  & 2 & 3 & 4 & 5 & 6 & 7\\ \cline{2-9}
        & \multirow{5}{*}{\rotatebox[origin=c]{90}{\shortstack{Number\\ of\\ robots\\ $n_R$}}} & 2 & $2.920$ & $6.550$ & $16.246$ & $47.311$ & $182.181$ & $915.320$\\
        \cline{3-9}
        & & 3 & * & $29.489$ & $96.035$ & $328.721$ & $1264.441$ & $5921.776$\\
        \cline{3-9}
        & & 4 & * & * & $376.697$ & $1600.171$ & $6785.22$ & $34364.22$\\
        \cline{3-9}
        & & 5 & * & * & * & $5692.486$ & $28992$ & $167579.7$\\
        \hline
    \end{tabular}}
\end{table}

\end{document}